\documentclass[conference,letterpaper]{IEEEtran}
\usepackage[margin=1in]{geometry}
\usepackage[utf8]{inputenc} % allow utf-8 input
\usepackage[T1]{fontenc}    % use 8-bit T1 fonts
\usepackage{hyperref}       % hyperlinks
\usepackage{amsfonts}       % blackboard math symbols
\usepackage{nicefrac}       % compact symbols for 1/2, etc.
\usepackage{microtype}      % microtypography
\usepackage{amsmath}
\usepackage{times}
\usepackage{amsthm}
\usepackage{graphicx}
\usepackage{algorithm}
\usepackage{algorithmic}
\usepackage{multirow}
\usepackage{booktabs}       % for professional tables
\usepackage{url}            % simple URL typesetting
\usepackage{mathtools}
\usepackage[caption=false]{subfig}
\usepackage{bm}
\usepackage{tikz}
\usepackage{cleveref}
\usetikzlibrary{bayesnet}
\usetikzlibrary{calc}
\newtheorem{proposition}{Proposition}

\title{Variational Federated Multi-Task Learning}
\author{\IEEEauthorblockN{Luca Corinzia, 
Ami Beuret, and Joachim M. Buhmann}
\IEEEauthorblockA{
Department of Computer Science\\
ETH Zürich, Switzerland\\
Email: \{luca.corinzia, ami.beuret, jbuhmann\}@inf.ethz.ch}}

\begin{document}
\maketitle

\begin{abstract}
In federated learning, a central server coordinates the training of a single model on a massively distributed network of devices. This setting can be naturally extended to a multi-task learning framework, to handle real-world federated datasets that typically show strong statistical heterogeneity among devices. Despite federated multi-task learning being shown to be an effective paradigm for real-world datasets, it has been applied only on convex models. In this work, we introduce {\tt VIRTUAL}, an algorithm for federated multi-task learning for general non-convex models. In {\tt VIRTUAL} the federated network of the server and the clients is treated as a star-shaped Bayesian network, and learning is performed on the network using approximated variational inference. We show that this method is effective on real-world federated datasets, outperforming the current state-of-the-art for federated learning, and concurrently allowing sparser gradient updates.
\end{abstract}

\section{Introduction}
\label{section:intro}
Large scale networks of remote devices, like mobile phones, wearables, smart homes, self-driving cars, and other IoT devices are becoming a significant source of data to train statistical models. 
As a consequence, there has been a growing interest to develop machine learning paradigms that can take into account distributed data-structure, despite the several challenges arising in this setting. 
(i) Security: Data generated by remote devices is often privacy-sensitive and its centralized collection and storage is governed by data protection regulations (e.g GDPR \cite{voigt2017eu} and the Consumer Privacy Bill of Rights \cite{house2012consumer}). 
Learning paradigms that do not access user data directly are hence desired. 
(2) System: Remote devices in these networks have typically important storage and computational capacity constraints, limiting the complexity and the size of the model that can be used. Moreover, the communication of information between devices or between the central server and devices, mostly happens on wireless networks. Hence communication cost can become a significant bottleneck of the learning process.
(3) Statistical: The devices of the network typically generate samples with different user-dependent probability distributions, making the setting in general strongly non-IID. 
While it is a challenge to achieve high statistical accuracy for classical federated and distributed algorithms in this setting, a multi-task learning (MTL) approach can tackle heterogeneous data more naturally.
Every device of the network requires a task-specific model, tailored for its own data distribution, to boost the performance of each task.

Federated learning (FL) \cite{mcmahan2017communication} has emerged to address the scenario of learning models on private distributed data sources. It assumes a federation of devices called \emph{clients} that both collect the data and execute an optimization routine, and a \emph{server} that coordinates the learning process by receiving and sending updates from and to the clients. This paradigm has been applied successfully in many real-world cases, e.g, to train smart keyboards in commercial mobile devices \cite{yang2018applied} and to train privacy-preserving recommendation systems \cite{ammad2019federated}. Federated Averaging ({\tt FedAvg}) \cite{mcmahan2017communication, nilsson2018performance} is the state-of-the-art for federated learning with \emph{non-convex} models and requires all clients to share the same model. Hence, it does not address the statistical challenge of strongly skewed data distributions, and while it has been shown to work well in practice for a range of (non-federated) real-world datasets, it performs poorly in heterogeneous scenarios \cite{mcmahan2017communication}. We address this problem by introducing {\tt VIRTUAL} (\textbf{V}ar\textbf{I}ational fede\textbf{R}aTed m\textbf{U}lti t\textbf{A}sk \textbf{L}earning), a new framework for federated MTL. In {\tt VIRTUAL}, the central server and the clients form a hierarchical Bayesian network and the inference is performed using variational methods. Every client has a task-specific model that benefits from the server model in a transfer learning fashion with \emph{lateral connections}. A part of the parameters are shared between all clients, and another part is private and tuned separately. The server maintains a posterior distribution that represents the plausibility of the shared parameters. In one step of the algorithm, the posterior is communicated to the clients before the training starts, while during training the clients update the posterior given the likelihood of their local data. Finally, the posterior update is sent back to the central server.
\paragraph{Contributions} Our main contributions are twofold: (i) We address for the first time the problem of federated MTL for \emph{generic non-convex models}, designing an additional MT metric and proposing {\tt VIRTUAL}, an algorithm to perform federated training with strongly non-IID client data distributions. (ii) We perform extensive experimental evaluation of {\tt VIRTUAL} on real-world federated datasets, showing that it outperforms the current state-of-the-art in FL, and simultaneously allowing lower communication costs.
\section{The {\tt VIRTUAL} algorithm}
In FL, $K$ clients are associated with $K$ datasets $\mathcal{D}_1,\dots,\mathcal{D}_K$, where $\mathcal{D}_i \coloneqq \{ \textbf{x}_i^{(n)},y_i^{(n)}\}_{n=1}^{N_i}$ is in general generated by a client dependent probability distribution function (pdf) and only accessible by the respective client. 
It is natural to fit $K$ different models, one for each dataset, enforcing a relationship between models using parameter sharing \cite{collobert2008unified}. 
This approach has been investigated extensively, and it has been shown to boost effective sample size and performance in MTL for neural networks \cite{ruder2017overview}.
\subsection{The Bayesian network}
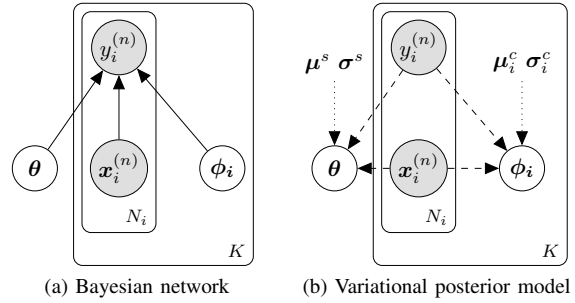
\begin{figure}
\centering
\subfloat[Bayesian network]{
\resizebox{!}{3.5cm}{
\begin{tikzpicture}
\node[obs]                               (x)         {$\bm{x}_i^{(n)}$};
\node[obs, above=of x]                   (y)         {$y_i^{(n)}$};
\node[latent, left=0.5cm of x]             (theta)     {$\bm{\theta}$};
\node[latent, right=0.8cm of x]            (phi)     {$\bm{\phi_i}$};
\edge {x,theta,phi}{y};
\plate {client} {(x)(y)} {$N_i$} ;
\plate {network} {(x)(y)(phi)(client.north west)(client.south east)} {$K$} ;
\end{tikzpicture}
}
\label{fig:bayesian_net}
}
\subfloat[Variational posterior model]{
\resizebox{!}{3.5cm}{
\begin{tikzpicture}
\node[obs]                               (x)         {$\bm{x}_i^{(n)}$};
\node[obs, above=of x]                   (y)         {$y_i^{(n)}$};
\node[latent, left=0.5cm of x]             (theta)     {$\bm{\theta}$};
\node[latent, right=0.8cm of x]            (phi)     {$\bm{\phi_i}$};
\node[draw=none,fill=none, above=of theta] (theta_var) {$\bm{\mu}^s$ $\bm{\sigma}^s$};
\node[draw=none,fill=none, above=of phi] (phi_var) {$\bm{\mu}^c_i$ $\bm{\sigma}^c_i$};
\edge[dashed] {x,y}{theta};
\edge[dashed] {x,y}{phi};
\edge[dotted] {phi_var}{phi};
\edge[dotted] {theta_var}{theta};
\plate {client} {(x)(y)} {$N_i$} ;
\plate {network} {(x)(y)(phi)(phi_var)(client.north west)(client.south east)} {$K$} ;
\end{tikzpicture}
}
\label{fig:bayesian_net_proxy}
}
\caption{Graphical models that describe the {\tt VIRTUAL} framework for federated learning. The plates represent replicates. In both figures, the outer plate replicates client $i$ over the total number of clients $K$, while the inner plate replicates sample index $n$ over the total number of samples per client $N_i$. Shadowed nodes represent observed variables and non-shadowed nodes represent latent variables.
(a) Solid lines denote the discriminative model $p(y_i^{(n)}| \bm{x}_i^{(n)}, \bm{\theta}, \bm{\phi}_i)$. (b) Graphical model of the approximated variational posterior. Dashed lines denote (deterministic) dependencies in the approximated variational posterior while dotted lines denote stochastic dependencies. Here we indicate as $(\bm{\mu}^s$,$\bm{\sigma}^s)$ and $(\bm{\mu}^c_i$,$\bm{\sigma}^c_i)$ the collection of all Gaussian parameters of server and client $i$.}
\end{figure}

Let assume a star-shaped Bayesian network with a server $S$ with model parameters $\bm{\theta}$, as well as $K$ clients with model parameters $\{\bm{\phi}_i\}_{i=1}^K$.
Assume that every client is a discriminative model distribution over the input given by $p(y_i^{(n)}| \bm{x}_i^{(n)}, \bm{\theta}, \bm{\phi}_i)$ (a straight-forward extension of the work could consider also generative models). 
Each dataset $\mathcal{D}_i$ has a likelihood that factorizes as $p(\mathcal{D}_i| \bm{\theta},\bm{\phi_i}) = \prod_{n=1}^{N_i}p(y_i^{(n)}| \bm{x}_i^{(n)},\bm{\theta}, \bm{\phi}_i)$. Following a Bayesian approach, we assume a prior distribution over all network parameters $p(\bm{\theta},\bm{\phi_1},\dots,\bm{\phi_K})$. 
The posterior distribution over all parameters, given \emph{all} datasets $\mathcal{D}_{1:K} \coloneqq \{\mathcal{D}_1,\dots,\mathcal{D}_k\}$ reads then
\begin{equation}
\label{eq:posterior}
p(\bm{\theta},\bm{\phi_1},\dots,\bm{\phi_K} | \mathcal{D}_{1:K}) \propto \frac{\prod_{i=1}^K p(\bm{\theta},\bm{\phi_i} | \mathcal{D}_i)}{p(\bm{\theta})^{K-1}} 
\end{equation}
where we enforce that client-data is conditionally independent given server and client parameters, 
$p(\mathcal{D}_{1:K}| \bm{\theta},\bm{\phi_1},\dots,\bm{\phi_K}) = \prod_{i=1}^K p(\mathcal{D}_i| \bm{\theta},\bm{\phi_i})$
, and a factorization of the prior as $p(\bm{\theta},\bm{\phi_1},\dots,\bm{\phi_K}) = p(\bm{\theta}) \prod_{i=1}^K p(\bm{\phi_i})$
. The Bayesian network is illustrated in \Cref{fig:bayesian_net}.
\subsection{The optimization procedure}
The posterior given in \Cref{eq:posterior} is in general intractable and hence we have to rely on an approximation inference scheme (e.g.\ variational inference, sampling, expectation propagation \cite{bishop2006pattern}). Here we propose an expectation propagation ({\tt EP}) like approximation algorithm \cite{minka2001expectation} that has been shown to be effective and to outperform other methods when applied in the continual learning (CL) setting \cite{bui2018partitioned,nguyen2017variational}. Let us denote the collection of all client parameters by $\bm{\phi} = (\bm{\phi}_1,\dots,\bm{\phi}_K)$. Then we define a \emph{proxy} posterior distribution that factorizes into a server and a client contribution for every client $i$ as
\begin{equation}
\label{eq:factorization}
q(\bm{\theta},\bm{\phi}) = \left(\prod_{i=1}^K s_i(\bm{\theta}) \right) \left(\prod_{i=1}^K c_i(\bm{\phi}_i) \right).
\end{equation}
The fully factorization of both server and client parameters allows us to perform a client update that is independent from other clients, and to perform a server update in the form of an aggregated posterior that preserves privacy. 

Given a factorization of this kind, the general {\tt EP} algorithm refines one factor at each step. It first computes a refined posterior distribution where the refining factor of the proxy is replaced with the respective factor in the true posterior distribution. It then performs the update minimizing the Kullback-Leibler (KL) divergence between the full proxy posterior distribution and the refined posterior. The optimization to be performed for our particular Bayesian network and factorization is given by the following.
\begin{proposition}
\label{prop:ep_free_energy}
Assuming that at step $t$ the factor $i$ is refined, then the proxy pdf $s_i^{(t)}(\bm{\theta})$ and $c^{(t)}_i(\bm{\phi}_i)$ are found minimizing the variational free energy function $\mathcal{L}_i \coloneqq \mathcal{L}(s_i(\bm{\theta}),c_i(\bm{\phi}_i))$, with
\begin{align}
\label{eq:free_energy}
\mathcal{L}_i &= D_{KL}\left(s_i(\bm{\theta}) \frac{s^{(t-1)}(\bm{\theta})}{s^{(t-1)}_i(\bm{\theta})}\bigg|\bigg| p(\bm{\theta})^{\frac{1}{K}}\frac{s^{(t-1)}(\bm{\theta})}{s^{(t-1)}_i(\bm{\theta})} \right) \nonumber\\
&\hspace{1cm} + D_{KL}\left(c_i(\bm{\phi}_i) || p(\bm{\phi}_i) \right) \\
&\hspace{1cm} -\mathbb{E}_{\substack{s^{(t)}(\bm{\theta})\\c_i(\bm{\phi}_i)}} \log p(\mathcal{D}_i|\bm{\theta},\bm{\phi}_i) \nonumber
\label{eq:free_energy}
\end{align}
where $s^{(t)}(\bm{\theta}) = s_{i}(\bm{\theta})\prod_{j \ne i}^K s_j^{(t-1)}(\bm{\theta})$ is the updated posterior over the server parameters.
\end{proposition}
We can see that the variational free energy in \Cref{eq:free_energy} decomposes naturally into two parts. The terms that involve the client parameters $c_i(\bm{\phi}_i)$ correspond to the standard variational free energy terms of \emph{Bayes by backprop} \cite{blundell2015weight}. Note that, except for the natural entropic complexity cost given by the second KL term, no additional regularization is applied on the client parameters, that can hence be trained efficiently and network agnostic. The terms that involve the server posterior are instead the likelihood cost and the first KL term. This regularization restricts the server to learn an overall posterior close to the so-called cavity distribution $\tilde{s}_i(\theta)^t = p(\theta)^{1/K}\prod_{j \ne i} s_j(\theta)^t$, obtained by the proxy posterior distribution replacing the current refining factor $i$ by the prior. This constraint effectively forces the server to progress in a CL fashion \cite{nguyen2017variational}, learning from new federated datasets and avoiding catastrophic forgetting of the ones already seen.

The free energy in \Cref{eq:free_energy} can be optimized using gradient descent and the reparametrization trick \cite{kingma2013auto}. For simplicity, we use a Gaussian mean-field approximation of the posterior, hence for server and client parameters, the factorization reads respectively $s_i(\bm{\theta}) = \prod_{d=1}^{D^s} \mathcal{N}(\theta_{d}| \mu_{id}^s,
\sigma_{id}^s)$ and $c_i(\bm{\phi}_i) = \prod_{d=1}^{D_i^c} \mathcal{N}(\phi_{id}| \mu_{id}^c,\sigma_{id}^c)$, where $D^s$ and $\{D_i^c\}_{i=0}^K$ are respectively the total number of parameters of the server and client networks. A depiction of the full graphical model of the approximated variational posterior is given in \Cref{fig:bayesian_net_proxy}.
The pseudo-code of {\tt VIRTUAL} is described in \Cref{alg:virtual}. The structure of the algorithm is equivalent to the FedAvg \cite{mcmahan2017communication} and the FedProx \cite{sahu2018convergence} algorithms. 
At each round, a subset of clients is selected and trained with the local free energy given in \Cref{eq:free_energy}. 
The client update is then computed as the ratio of the client parameter distribution before and after the training. The ratio corresponds to the simple difference of the sufficient statistics in the case of exponential family distributions (see \Cref{app:gaussian_case}), that hence is equivalent to the delta computation in the FedAvg setting. In the general setting, it can be computed as the un-normalized pdf $\Delta_i = \frac{s_i^{(t)}(\bm{\theta})}{s_i^{(t-1)}(\bm{\theta})}$. 
The client $i$ communicates the delta $\Delta_i$ to the main server, that aggregates all the received updates into a single update $\Delta = \prod_{i \in \mathcal{C}_t} \Delta_i$. 
A major difference with the typical non-MTL setting, where the new server model is given by averaging all selected active clients at any given round, in our case we aggregate \emph{updates} and the information of non-active clients is effectively retained in the server posterior. 
In the simple case of mean-field Gaussian approximation of the posterior distribution, the delta computation and the aggregation easily generalize the subtraction and average used in FedAvg, taking into account the uncertainty in the parameters represented by the standard deviation of the learned Gaussian (see \Cref{app:gaussian_case}). Notice that similarly to {\tt FedAvg}, privacy is preserved since at any time the server can get access only to the overall posterior distribution $s(\bm{\theta})$ and to the aggregated update $\Delta(\bm{\theta})$, and never to the individual factor $s_i(\bm{\theta})$ and $c_i(\bm{\theta})$, that are visible only to the respective client.
\begin{algorithm}
\caption{{\tt VIRTUAL}}
\label{alg:virtual}
\begin{algorithmic}[1]
\STATE {\bfseries Input:} datasets $\{\mathcal{D}_1,\dots,\mathcal{D}_k\}$, $T$ number of rounds, priors $p(\bm{\theta}), \{p(\bm{\phi}_i)\}_{i=1}^K$, $C$ number of refined clients per round, $E$ number of training epochs per round.
\STATE initialize all pdfs $c_i^{(0)}(\bm{\phi}_i)$ and $s_i^{(0)}(\bm{\theta})$
\STATE $s^{(0)}(\bm{\theta}) \gets \prod_i s_i^{(0)}(\bm{\theta})$
\FOR{ round $t = 1,2 \dots, T$
}
    \STATE choose randomly a set of $C$ active clients $\mathcal{C}_t$ to be refined.
    \STATE (client) receives $s^{(t)}$ from server.
    \STATE (\emph{client}) compute new server prior $p(\bm{\theta})^{\frac{1}{K}}\frac{s^{(t)}(\bm{\theta})}{s^{(t-1)}_i(\bm{\theta})}$
    \STATE(\emph{client}) $s_i^{(t)}(\bm{\theta}),c^{(t)}_i(\bm{\phi}_i) \gets$ joint optimization of \cref{eq:free_energy} for $E$ epochs.
    \STATE (\emph{client}) compute delta $\Delta_i^{(t)}(\bm{\theta}) \gets \frac{s_i^{(t)}(\bm{\theta})}{s_i^{(t-1)}(\bm{\theta})}$
    \STATE (aggregate) $\Delta (\bm{\theta}) \gets \prod_{i \in \mathcal{C}_t} \Delta_i (\bm{\theta})$.
    \STATE (server) receives $\Delta (\bm{\theta})$ and applies it as $s^{(t+1)}(\bm{\theta}) \gets s^{(t)}(\bm{\theta}) \Delta(\bm{\theta}) $ to the server
\ENDFOR
\end{algorithmic}
\end{algorithm}
We can further notice an interesting similarity of the {\tt VIRTUAL} algorithm to the \emph{Progress\&Compress} method for CL introduced in \cite{schwarz2018progress}, where a similar free energy is obtained heuristically by composing CL regularization terms and distillation cost functions \cite{hinton2015distilling}.

In the experimental section we will make use of a slight modification of the free energy given in \Cref{eq:free_energy} where the KL divergence terms are weighted by a regularization multiplier $\beta$. The Kl multiplier has been widely used already in other scenarios, e.g., disentanglement in unsupervised learning \cite{Higgins2017betaVAELB}, where it has been shown that at different values of $\beta$ achieves various degree of disentanglement in the embedding space. We will show also in this case that a tuning of the KL multiplier can enhance the performance of the model.
\section{Related Work}
We here provide a brief survey of work in the area of distributed/federated learning and of transfer/continual learning, in light of the problem at hand described in \Cref{section:intro} and of the tools used in deriving {\tt VIRTUAL}.
\paragraph{Distributed and Federated Learning} Distributed learning is a learning paradigm for which the optimization of a generic model is distributed in a parallel computing environment with centralized data \cite{mcdonald2010distributed}. Early work on this paradigm propose various learning strategies that require iterative averaging of locally trained models, typically using Stochastic Gradient Descent ({\tt SGD}) steps in the local optimization routine \cite{mcdonald2010distributed,povey2014parallel, zhang2015deep,dean2012large}. 
Distributed learning typically consider the learning to be set in a computational cluster, hence with few computing devices, fast and reliable communication between devices, and centralized unbalanced datasets.
FL \cite{mcmahan2017communication} eliminates all these latter constraints and it is framed as a paradigm that encompasses the new challenges and desiderata listed in \Cref{section:intro}.
{\tt FedAvg} \cite{mcmahan2017communication, konevcny2016federated} has been proposed as a straightforward heuristic for the FL. At each step of the algorithm, a subset of online clients is selected randomly, and these are then updated locally using {\tt SGD}. The models are then averaged to form the model at the next step, which is maintained in the server and transmitted back to all clients. Despite working well in practice, it has been shown that the performance of {\tt FedAvg} can degrade significantly for skewed non-IID data \cite{mcmahan2017communication, zhao2018federated}. 

Some heuristics have been proposed recently to solve the statistical challenges of FL. In particular recently it has been proposed to share part of the client-generated data \cite{zhao2018federated} or a server-trained generative model \cite{jeong2018communication} to the whole network of clients. 
These solutions are however questionable since they require significant communication effort and do not comply with the standard privacy requirements of FL. 
Another solution for this problem has been proposed in \cite{sahu2018convergence}, where the authors extend {\tt FedAvg} into {\tt FedProx}, an algorithm that prescribes clients to optimize the local loss function, further regularized with an quadratic penalty anchored on the weights of the previous step. 
Despite showing improvements on the {\tt FedAvg} algorithm for highly data-heterogeneous settings, the method is strongly inspired by early works on continual and transfer learning (see, e.g., Elastic Weight consolidation in \cite{kirkpatrick2017overcoming,zenke2017continual} and the literature review in the next paragraph) and hence can be further refined. 

The first contribution to highlight the possibility of naturally embedding FL in the MTL framework has been reported by {\tt MOCHA} \cite{smith2017federated}, that extends some early work on distributed MTL-like {\tt CoCoA} and variations \cite{smith2018cocoa,jaggi2014communication,ma2015adding}. 
In this work a federated primal-dual optimization algorithm is derived for \emph{convex} models with MTL regularization, and it is shown for the first time that the MTL framework can enhance the model performance, with the MTL model outperforming global models (trained with centralized data) and local models as well, on real world federated datasets. 
This method, however, can be only used on convex models, hence it does not constitute a usable benchmark for deep learning models that are used in the experimental section here.

More recently, multiple efforts \cite{yurochkin2019bayesian, Wang2020Federated, singh2020model} have been made to align and match the weights of client models before aggregation in a layer-wise fashion. This is done to minimize the effect of averaging model weights that do not correspond to each other, due to the overparametrization and the symmetry of neural network parametrization. Note that, despite such efforts are applied to standard algorithms as FedAvg, they can be extended to our framework seamlessly.

A Bayesian approach for distributed datasets, similar to the method proposed in this paper, is developed in \cite{hasenclever2017distributed}, where Expectation Propagation (EP) \cite{minka2001expectation} and its variations are performed on a generic partition of the dataset for distributed inference. However, the authors propose only a \emph{global} model, hence with no structure of shared and non shared parameters between the server and clients. The training is further performed according to a single loss function, hence not in a MTL setting. Moreover, inference is performed using heavy MCMC methods to estimate the moment of the local distributions, limiting the scale of the model considered.
In \cite{vehtari2020expectation, bui2018partitioned}, a further variational framework for generic partitioned data is described. It can be noted that the frameworks proposed encompasse also our method if applied to the particular MTL BayesNet in \Cref{fig:bayesian_net}. However the case study and the experimental section are focused on a classic EP algorithm based on moment matching and heavy MCMC simulation for moment estimation, and hence it can only address limited size models.

\paragraph{Transfer and Continual Learning} The transfer of knowledge in neural network, from one task to another, has been used extensively and with great success since the pioneering work in \cite{hinton2006reducing} of transferring information from a generative to a discriminative model using fine tuning. 
The application of this straightforward procedure is however difficult to apply in scenarios where multiple tasks from which to transfer from are available. 
Indeed a good target performance can be obtained only with a priori knowledge of task similarity, that is usually not known, while learning of sequential tasks causes knowledge of previous tasks to be abruptly erased from the network in what has been called \emph{catastrophic forgetting} \cite{french1999catastrophic}. 

Many methods have been introduced to overcome catastrophic forgetting, and to enable models to learn multiple task sequentially retaining a good overall performance, and transferring effectively to new tasks. Many early works proposed different regularization terms of the loss function anchored to the previous solution in order to achieve new solutions that generalize well on old tasks \cite{kirkpatrick2017overcoming,zenke2017continual}. These methods have been first introduced as heuristics, but have been found to be applications of well-known inference algorithms like \emph{Laplace Propagation} \cite{smola2003laplace} and \emph{Streaming Variational Bayes} \cite{broderick2013streaming}, which led to further generalizations \cite{lee2017overcoming,geyer2019transfer}. New approaches focused on other components, like architecture innovations, introducing lateral connections that allow new models to reuse knowledge from previously trained models with \emph{layer-wise adaptors} \cite{rusu2016progressive,schwarz2018progress}, and memory enhanced models with generative networks \cite{wu2018memory,yoon2019oracle}. A recently introduced online Bayesian inference approach \cite{nguyen2017variational} served as inspiration for our work. It frames the continual learning paradigm in the Bayesian inference framework, establishing a posterior distribution over network parameters that is updated for any new task in light of the new likelihood function. It has been shown that this method outperformed all previously known methods for CL.
\section{Experiments}
In this section we present an empirical evaluation of the performance of {\tt VIRTUAL} on three real-world federated datasets that well represent both the challenges of federated training and of multi-task learning. Due to the fact that in VIRTUAL clients retain a private model at every round, experiments performed on a simulated network on a single GPU have a memory cost that scales linearly with the number of clients (compared to FedAvg that has a constant memory cost on simulation as well). For this reason, the number of clients is bounded in all experiments to 100. Note however that this drawback does \emph{not} extend to a real network of devices, as in the latter case the client model is retained by the device, and the memory cost at the server side is constant w.r.t the number of clients.

\subsection{Dataset description}
\begin{table}[ht]
\caption{Statistics of the datasets used in the experiments.}
\label{tab:dataset}
\centering
\begin{tabular}{@{}lrrrr@{}} 
\toprule
\textbf{Dataset} &\textbf{K} &\textbf{Size} ($\approx$) & \multicolumn{2}{c}{\textbf{Size/K}}\\
\cmidrule{4-5}
&  & & mean & std \\
\midrule
FEMNIST & 100 & 55k & 0.5k & 54\\
MNIST & 100 & 60k & 600 & 0\\
PMNIST & 100 & 60k & 600 & 0 \\
VSN  & 23 & 68k & 3k & 559 \\
HAR & 30 & 15k & 0.5k & 56\\
NLP & 100 & 1.2m & 13k & 11k\\
\bottomrule
\end{tabular}
\end{table}
\par \textbf{FEMNIST}: This dataset consists of a federated version of the EMNIST dataset \cite{cohen2017emnist}, maintained by the \emph{LEAF} project \cite{caldas2018leaf}. Different clients correspond to different writers. We subsample 100 random writers and use only the 10 digit labels. Train and test split is provided by the distribution.
\par \textbf{Vehicle Sensors Network (VSN)}\footnote{\url{http://www.ecs.umass.edu/~mduarte/Software.html}}: A network of 23 different sensors (including seismic, acoustic, and passive infra-red sensors) are place around a road segment to classify vehicles driving through. \cite{duarte2004vehicle}. The raw signal is featurized in the original paper into 50 acoustic and 50 seismic features. We consider every sensor as a client and perform the binary classification of assault amphibious and dragon wagon vehicles.
\par \textbf{Human Activity Recognition (HAR)}\footnote{\url{https://archive.ics.uci.edu/ml/datasets/Human+Activity+Recognition+Using+Smartphones}}: Recordings of 30 subjects performing daily activities are collected using a waist-mounted smartphone with inertial sensors. The raw signal is divided into windows and featurized into a 561-length vector \cite{anguita2013public}. Every individual corresponds to a different client and we perform classification of 12 different activities (e.g., sitting, walking). For both the VSN and the HAR, a 75\%-25\% train-test split is performed.
\par \textbf{MNIST}: The classic MNIST dataset \cite{lecun1998gradient}, randomly split into 100 different sections, one section per client. Every client has 600 training samples and 100 test samples. This dataset represents an atypical federated dataset with very homogeneous clients, both in terms of dataset sizes and in term of statistical properties of samples.

\textbf{Permuted MNIST (PMNIST)}: The MNIST dataset is randomly split into 100 sections as above, and a random permutation of pixels is applied to every single client dataset. This dataset has been introduced in \cite{goodfellow2013empirical} in the context of continual learning and represent a strongly non-IID federated dataset, with low level features being very dissimilar between clients.

\textbf{Shakespeare (NLP)}: This dataset is built concatenating the whole literary production of William Shakespeare \cite{karpathy2015unreasonable}. The task here considered is English words spelling, hence the next character prediction task, over a vocabulary size of 86. The characters are arranged in sequences of 80 and further aggregated in batches of 10 sequences. Every role of a play is considered as a individual client, and we excluded roles that do not contain a single full batch.
\par A comprehensive description of the statistics of the datasets used is available in \Cref{tab:dataset}.

\subsection{Experimental setting} 
We consider a multilayer perceptrons (MLP) with two hidden dense layers (with local reparametrization for the Bayesian counterpart \cite{kingma2015variational}) with 100 units and ReLU activation functions in the hidden layers, and softmax activation at the output layer. 
For the NLP task, we use a two-layer LSTM classifier with 100 hidden units per layer and an 8D embedding layer. The Bayesian model makes use of Bayesian LSTMs \cite{fortunato2017bayesian} and Bayesian Gaussian Embeddings \cite{vilnis2015word}.
We further use a convolutional neural network for the FEMNIST dataset with two convolutional layers with kernel size 5 and number of filter respectively 32 and 64. We use max-pooling after both convolutional layers, then we adopt an MLP as described above on the flattened activations.

Using the notation of the original paper \cite{mcmahan2017communication}, all methods are evaluated in all experiments with a number of updated clients per round $C=10$ and a number of epochs per round $E=20$, that is high enough to be meaningful on a FL setting, to guarantee convergence in all scenarios \cite{mcmahan2017communication,caldas2018leaf} while being challenging on complex tasks, like in the case of the Shakespeare dataset. 
Hyperparameter optimization is performed over a grid of 5 log-spaced \emph{client} learning rates. Vanilla SGD optimizer is used as the local optimizer for every client. 
Implementation of {\tt VIRTUAL} is based on \emph{tensorflow} \cite{abadi2016tensorflow} and \emph{tensorflow distributions} \cite{dillon2017tensorflow} packages. 

\subsection{Metrics} 
At every round and for any given metric (cross-entropy loss or accuracy), we report two different variations: (i) The centralized metric measured by the central server (S) tested on all client test data; (ii) the Multi-Task (MT) metric measured as the average test metric of all clients models. 
In both cases, the average is weighted by the client dataset size. For the case of a non-MTL approach (like FedAvg and FedProx), we measure the MT metric using the model that every client deployed last, while in the case of an MTL setting as in Virtual, every client maintains a private model that is tested at every round.

\subsection{Effect of the kl divergence weight}
\begin{figure}
\centering
\includegraphics[width=1.\columnwidth]{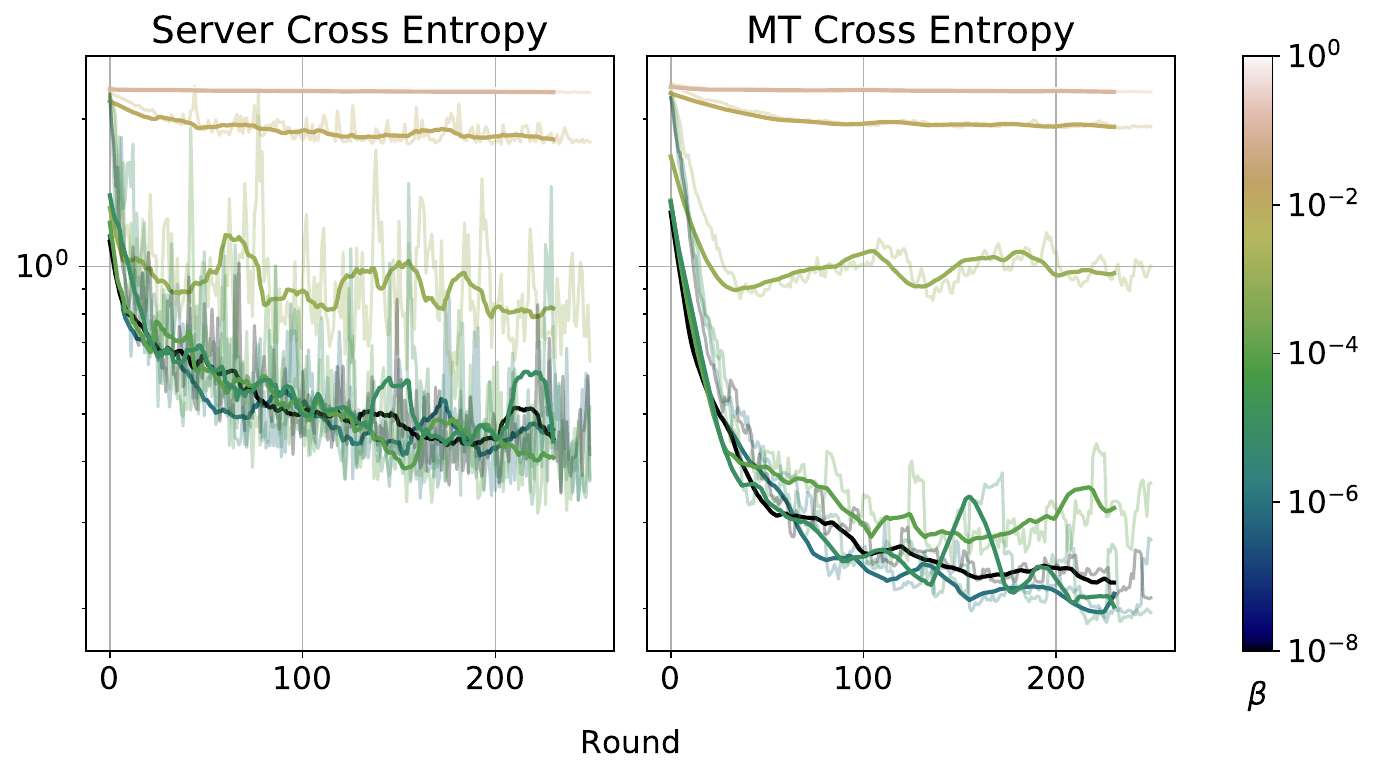}
\caption{Regularization effect of the KL divergence on the cross-entropy loss. For the FEMNIST dataset, we report the server and the MT cross-entropy loss during training at different values of the KL divergence multiplier $\beta$. Thick lines represent moving averages with window size 20. Log-scale is applied on both the y scale and the color bar.}
\label{fig:kl_contribution}
\end{figure}

We first study the effect of the KL divergence multiplier $\beta$ on the performance of an MLP network in the FEMNIST dataset. 
We can observe from \Cref{fig:kl_contribution} that values of $\beta$ in the range $10^{-6}-10^{-3}$ do not impair the performance of the model. For a higher value of $\beta$ the leading term in the free energy \Cref{eq:free_energy} is given by the regularization term and the reconstruction loss is negligible. 
From \Cref{fig:kl_contribution} we can further observe that in the range of adequate values of $\beta$, the server performance is not affected, while increased generalization is achieved in the MT loss, with the $\beta=10^{-5}$ being the best performing model. 
In all following experiments, we tune the $\beta$ parameter and the $l_2$ regularizer in the case of the FedProx baseline (with $\beta = 0$ corresponding to FedAvg) in a log-spaced grid of 5 values. 
We denote as FedProx the best model with a $l_2$ multiplier strictly larger than 0.
\begin{figure*}[ht]
\centering
\includegraphics[width=1.\textwidth]{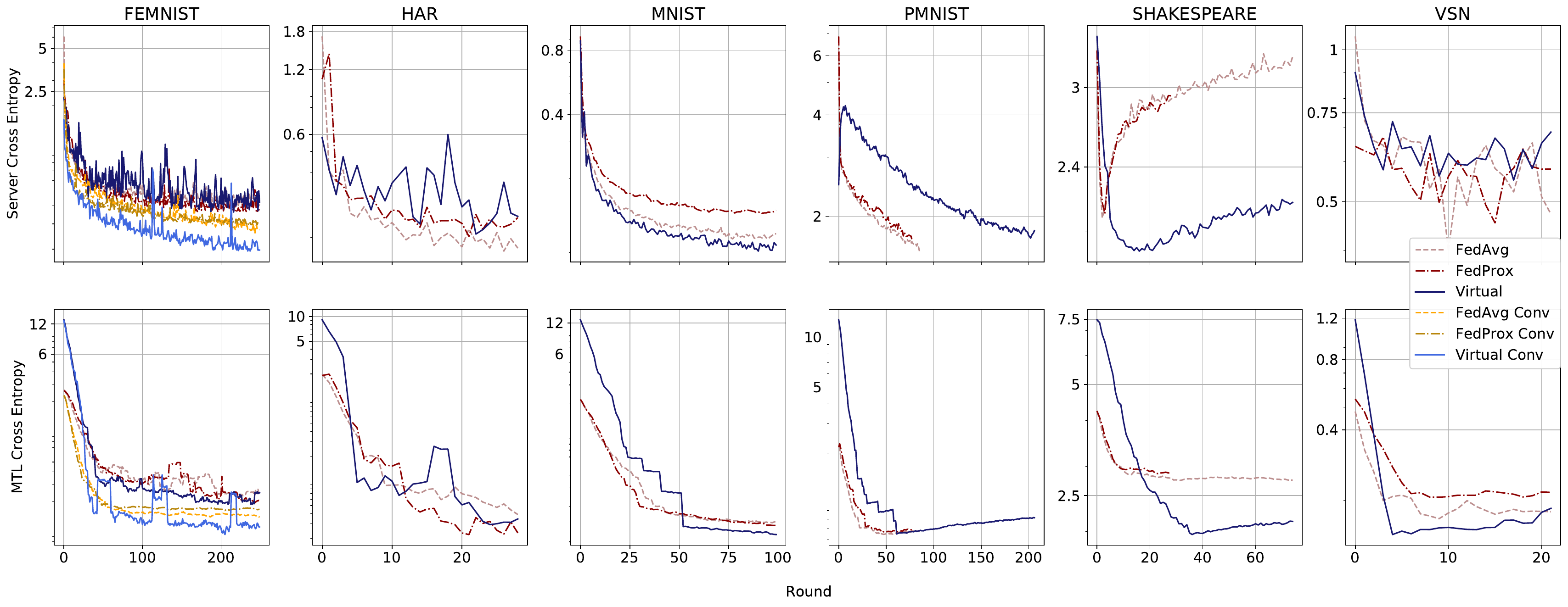}
\centering
\caption{Central server (first row) and multi-task (MT, second row) cross-entropy loss as a function of the federated training round. For every dataset (column) we report the loss of the two baselines FedAvg and FedProx and our method Virtual. For the FEMNIST dataset, we report both the performance of the MLP and the convolutional NN architecture. For all the other datasets, only the performance of one model is showed (LSTM architecture in the Shakespeare datasets, MLPs on all the others). Log-scale is used in the y-axis.}
\label{fig:results}
\end{figure*}

\subsection{Full results}
In this section, we evaluate the performance of our proposed method and the baselines FedAvg and FedProx in terms of both server and multi-tasks performance. 

\textbf{Metric comparison.} 
From \Cref{fig:results} we can first observe that the MT metric is uniformly more stable than the respective server variant since the former assumes clients retaining a private model until further training, while the latter assumes a server model that is updated at every single round, hence experiencing more stochasticity during the training process. Moreover, the MT metric is typically delayed compared to the server counterpart since clients are updated on average only every $K/C$ rounds. This is responsible for the slower progress of the MT metric, with the values of the losses in the first stages of training being up to a factor of 10 larger than the server counterpart. 
We further notice from \Cref{fig:results} (last column) that the MT metrics are typically superior to the centralized metrics (with some exceptions that are examined thoroughly in the following), showing that, at convergence, clients can personalize the model to a specific private dataset.

\textbf{Method comparison.}
In \Cref{tab:results} we additionally report the maximum accuracy achieved at convergence by the baselines FedAvg and FedProx and our method in all datasets. 
We can observe that our method outperforms both baselines in almost all datasets (except in the PMNIST datasets) and with all neural network architectures considered (MLPs, convolutional and RNNs). 
Virtual is able to achieve up to $+2\%$ and $+1\%$ in maximum accuracy respectively in the MT and S variant in the FEMNIST, MNIST and Shakespeare datasets, and marginally outperforms the baselines in VSN. 
The Shakespeare dataset is particularly crucial. 
In this dataset, the S metrics are superior then the MT metrics, implying that, at convergence, the clients that train further on the private datasets impair their performance. 
This follows likely from the high heterogeneity of the size of the Shakespeare datasets (see the last row in \cref{tab:dataset}), that can cause clients with a small dataset to over-fit the private data and under-perform the central server model. 
Our proposed method performs particularly well in this scenario, causing only a slight reduction of the MT compared to the S accuracy.
\begin{table}[ht]
\caption{Server (S) and Multi-Task (MT) max accuracy over all dataset at convergence. Values are given in percentage.}
\label{tab:results}
\centering
\begin{tabular}{@{}llccc@{}} 
\toprule
\textbf{Dataset} & \textbf{Metric} &
\textbf{FedAvg} & \textbf{FedProx} & \textbf{Virtual} \\
\midrule
\multirow{2}{*}{\shortstack[l]{FEMNIST\\ MLP}} & MT & 94.3 & 94.5 & \bfseries 95.7 \\
& S &  90.2 & 89.9 & \bfseries 90.9 \\
\multirow{2}{*}{\shortstack[l]{FEMNIST\\ Conv}} & MT & 97.3 & 97.0 & \bfseries 98.2 \\
& S & 95.5 & 95.5 & \bfseries 97.3 \\
\multirow{2}{*}{MNIST} & MT & 96.9 & 96.9 & \bfseries 97.4 \\
& S & 97.6 & 97.6 & \bfseries 97.8 \\
\multirow{2}{*}{PMNIST} & MT & \bfseries 85.9 & 85.6 & 84.2 \\
& S & \bfseries 48.3 & 47.5 & 42.5\\
\multirow{2}{*}{VSN} & MT & 96.2 & 96.0 & \bfseries 96.6 \\
& S & \bfseries 89.6 & 89.5 & 88.8\\
\multirow{2}{*}{HAR} & MT & 98.9 & \bfseries 99.4 & 99.1  \\
& S & 94.0 & 94.0 & \bfseries 94.3 \\
\multirow{2}{*}{NLP} & MT & 46.1 & 46.4 & \bfseries 48.4 \\
& S & 48.1 & 48.5 & \bfseries 48.6 \\
\bottomrule
\end{tabular}
\end{table}

\subsection{Inducing sparse updates}
\begin{figure}[ht]
\centering
\includegraphics[width=\columnwidth]{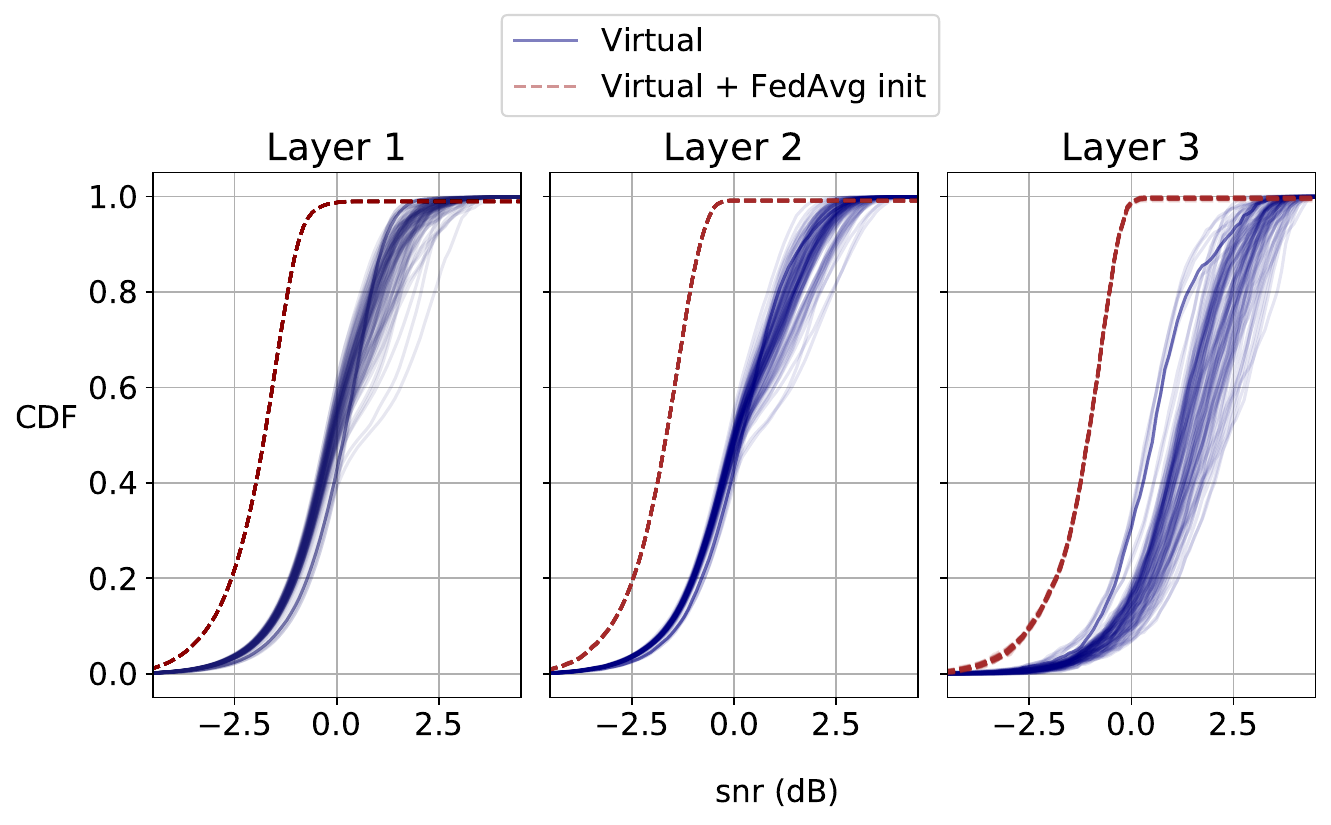}	
\centering
\caption{Cumulative distribution function of the signal-to-noise ratio (reported in log scale) for all clients and for the three consecutive dense layer of the network. Clients that are not initialized with the server weights at each round show a more compressible model. Simulation performed on the FEMNIST dataset on an MLP architecture.}
\label{fig:snr}
\end{figure}

The use of a Bayesian framework allows us to estimate the importance of the client $i$ using the property of the client posterior distributions and client un-normalized updates distributions $\Delta_i$. 
In \cref{fig:snr} we examine the cumulative distribution function (CDF) of the signal-to-noise (SNR) of the weights of the client posteriors. A CDF located on the right-hand side of the plot represents a compressible network, with only a small ratio of the weights with high SNR being determinant in the model performance (see \cite{blundell2015weight} for an application of this concept to simple neural network pruning).
For comparison, we also implement a variation of the Virtual method that retains the client loss function used in Virtual, and in addition initializes the weights of the client with the server posterior at the beginning of every training round (Virtual + FedAvg init in \cref{fig:snr}). In these plots, we observe that the server initialization forces all clients to learn a complex model (larger CDF), while with no such initialization the clients specialize to the task at hand with a small ratio of weights.
We, therefore, design a simple updated pruning procedure that sparsifies the client updates setting to zero all $\Delta_i$ elements that have a SNR smaller than a given percentile. 
The results are reported in \cref{tab:sparse_update}.
Virtual greatly outperforms the FedAvg initialization variant and retains a superior performance compared to the FedProx method (compare with \Cref{tab:results}, first and second row) up to an induced sparsity of 75\%.
This is equivalent to a 50\% communication reduction cost compared to FedProx, factoring the use of a Bayesian Neural Network that uses twice as many parameters as a standard deterministic network.

\begin{table}[ht]
\caption{Max accuracy at various levels of updates sparsity. Pruning is performed using the signal-to-noise-ratio of the updates.}
\label{tab:sparse_update}
\centering
\begin{tabular}{@{}cllcc@{}} 
\toprule
\multicolumn{1}{p{1.7cm}}{\centering \textbf{\% pruned \\ weights}} &
\multicolumn{1}{p{0.9cm}}{\centering \textbf{Delta \\ size}}
& \textbf{Acc} & \multicolumn{1}{p{2.1cm}}{\centering \textbf{Virtual +\\ FedAvg init}} & \textbf{Virtual} \\
\midrule
\multirow{2}{*}{0\%} & \multirow{2}{*}{158k} & MT & 86.2 & \bfseries 95.6 \\
& & S & 89.4 & \bfseries 90.9  \\
\multirow{2}{*}{50\%} & \multirow{2}{*}{79k} & MT & 76.3  & \bfseries 94.3\\
& & S & 81.8 & \bfseries 89.2 \\
\multirow{2}{*}{75\%} & \multirow{2}{*}{40k} & MT & 78.7 & \bfseries 94.9 \\
& & S & 80.4 & \bfseries 90.8 \\
\multirow{2}{*}{90\%} & \multirow{2}{*}{16k} & MT & 25.5 & \bfseries 48.5 \\
& & S & \bfseries 21.0 & 20.3 \\
\bottomrule
\end{tabular}
\end{table}
Note that the scenario here described is very different from usual pruning performed on NNs, as in the latter case the pruning is performed only at test time (or after fine tuning) on fully trained NN \cite{han2015learning}, or a priori on a carefully initialized NN \cite{frankle2018lottery} while in our  case the pruning is performed at every training round (hence at every E epochs of training), making it much a much harder task at a high level of pruning.
\section{Conclusion}
In this work we introduced {\tt VIRTUAL}, an algorithm for federated learning that tackles the well-known statistical challenges of the federated learning framework in a multi-task setting. 
We consider the federation of a central server and clients as a Bayesian network and perform training using approximated variational inference. 
The algorithm naturally complies with the federated setting desiderata, giving access to the central server only to an aggregated parameter update in the form of an overall posterior distribution over shared parameters. 
The algorithm is shown to outperform the state-of-the-art in many IID and non-IID real world federated datasets.

One possible direction for further developments is to consider synchronous updates of multiple clients (as preliminary seen already in \cite{bui2018partitioned}) studying empirically the effect of using outdated priors during client training or theoretically developing a new Bayesian model of synchronous updates.

Another interesting direction is the exploration of other design choices. Indeed the general method can be tuned for a particular application by modifying, e.g., the architecture of lateral connections between devices (Block-Modular NN \cite{terekhov2015knowledge}, NinN architecture \cite{lin2013network}), the topology of the Bayesian network (star shape, hierarchical etc.), the choice of the variational inference algorithm. 
Finally, it is possible to study thoroughly {\tt VIRTUAL} under memory constraints, using more sophisticated pruning procedures, sparsity inducing losses or using  optimal strategy for data storage, in the line of coresets theory.

\clearpage

\appendices
\crefalias{section}{appendix}

\section{Additional experiments}
\label{sec:additional_experiments}
\begin{figure*}[ht]
\centering
{
\subfloat[Training cross entropy loss as a function of the federated training round. Log-scale is used in the y-axis.]{
\label{fig:losses_20}
\includegraphics[width=0.97\textwidth]{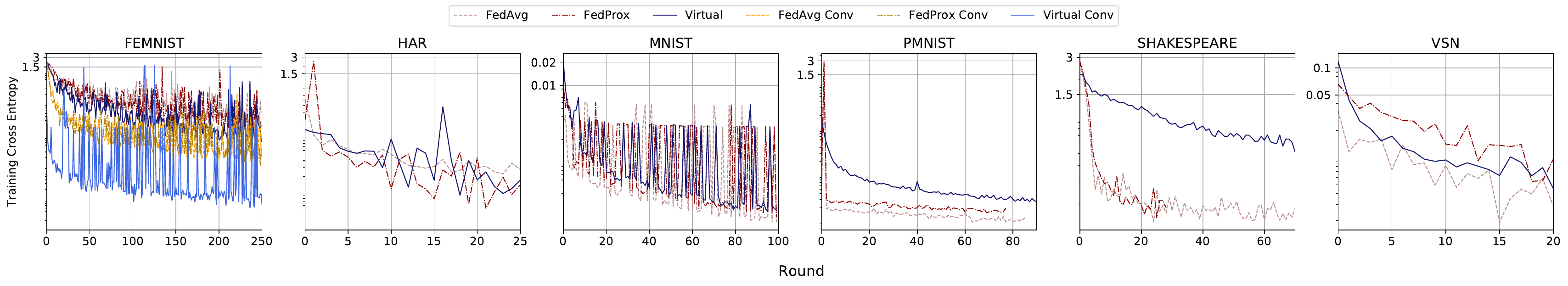}
}
\\
\subfloat[Server, Multi-Task and training accuracy over all datasets considered.]{
\label{fig:accuracies_20}
\includegraphics[width=0.97\textwidth]{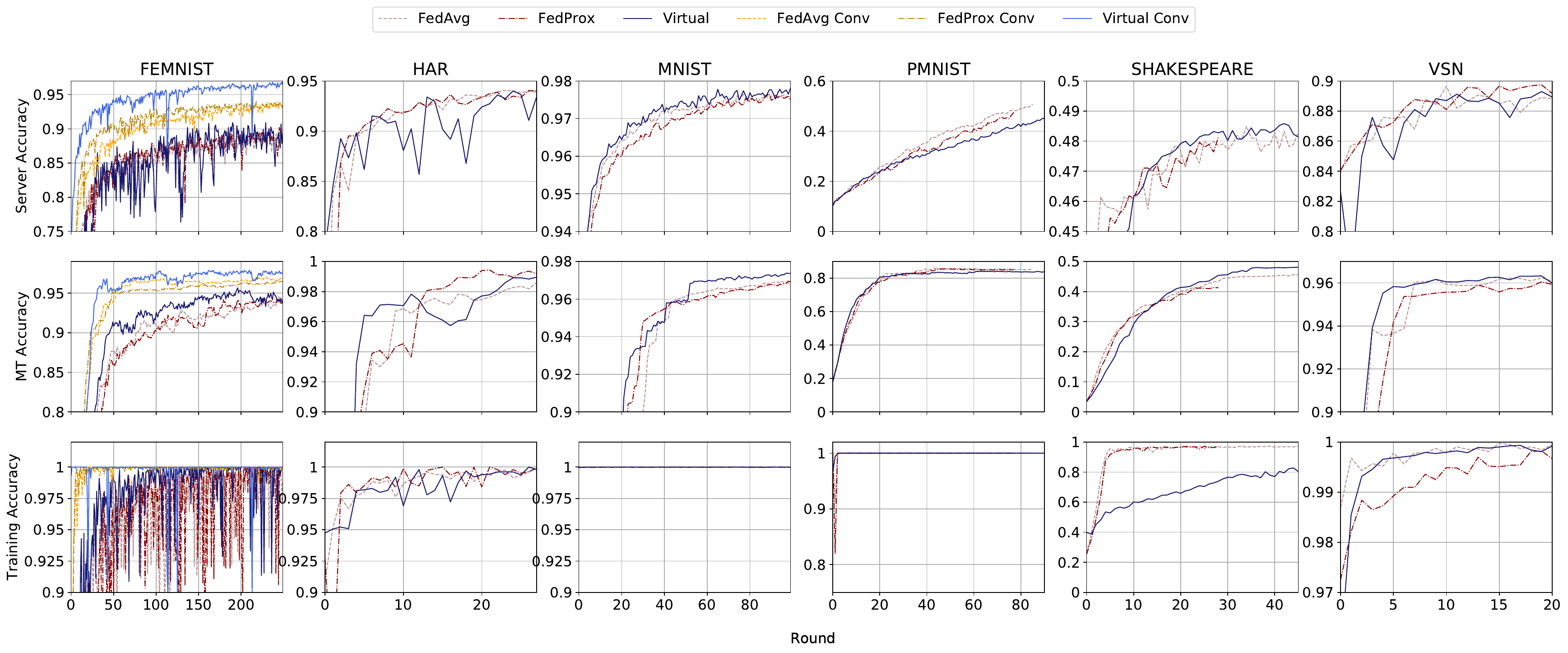}
}
}
\caption{Additional learning curves.}
\end{figure*}
In this section, we report the additional learning curves that are not reported in the main text, i.e., training cross-entropy in \Cref{fig:losses_20} and the server, MT and training accuracies in \Cref{fig:accuracies_20} for the dataset considered, and for both MLPs and ConvNet architectures in the case of the FEMNIST dataset.

\begin{figure*}[ht]
\centering
\includegraphics[width=\textwidth]{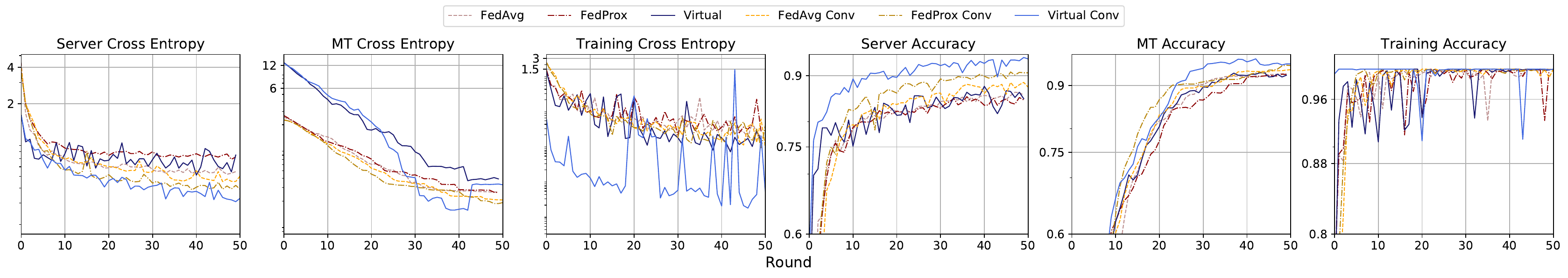}
\caption{Learning curves for the FEMNIST dataset with epochs per round set at $E=100$, simulating a high node computation scenario.}
\label{fig:losses_100}
\end{figure*}

In \Cref{fig:losses_100} we also report the learning curves obtained in an additional experiment performed on the FEMNIST dataset, with the epochs per round parameter set to $E=100$, to show the applicability of the method in scenarios of higher node computational load. We can observe that also in this scenario Virtual outperforms both FedAvg and FedProx using MLPs and ConvNet architectures. In \cref{tab:results_100} the maximum MT and server accuracies are reported, with Virtual achieving a +0.4\% and +0.9\% respectively on MLPs and convolutional architectures. 

\begin{table}[ht]
\caption{Server (S) and Multi-Task (MT) maximum accuracy over all datasets at convergence, with $E=100$ epochs per round. Values are given in percentage.}
\label{tab:results_100}
\centering
\begin{tabular}{@{}llccc@{}} 
\toprule
\textbf{Dataset} & \textbf{Metric} &
\textbf{FedAvg} & \textbf{FedProx} & \textbf{Virtual} \\
\midrule
\multirow{2}{*}{\shortstack[l]{FEMNIST\\ MLP}} & MT & 92.7 & 92.8 & \bfseries 93.2 \\
& S & 86.2 & 86.6  & \bfseries 87.6\\
\multirow{2}{*}{\shortstack[l]{FEMNIST\\ Conv}} & MT & 97.0 & 97.3 & \bfseries 98.2 \\
& S & 95.5 & 95.6 & \bfseries 97.3 \\
\bottomrule
\end{tabular}
\end{table}

\section{The Gaussian case} 
\label{app:gaussian_case}
The Virtual method requires to compute the so-called (un-normalized) cavity distributions $\frac{s(\theta)}{s_i(\theta)}$, the client deltas $\Delta_i = \frac{s(\theta)}{s_i(\theta)}$, and deltas aggregation $\Delta = \prod_{i \in \mathcal{C}_t} \Delta_i$ (see Algorithm 1 in the main text). Using a factorized Gaussian distribution over the weights as $s_i(\bm{\theta}) = \prod_{d=1}^{D^s} \mathcal{N}(\theta_{d}| \mu_{id}^s,
\sigma_{id}^s)$, we can easily observe that the factorization extends to all three terms listed above. In turn, in order to implement the Virtual algorithm with factorized Gaussian distributions, we need to compute univariate Gaussian products and ratios that read respectively (see \cite[Sec 8.1.8]{petersen2008matrix}):
\begin{align*}
\mathcal{N}(x | \mu_1, \sigma_1^2) \cdot \mathcal{N}(x | \mu_2, \sigma_2^2) = \frac{1}{Z_p} \mathcal{N}(x | \mu_p, \sigma_p^2) \\
\frac{\mathcal{N}(x | \mu_1, \sigma_1^2)}{\mathcal{N}(x | \mu_2, \sigma_2^2)} = \frac{1}{Z_r} \mathcal{N}(x | \mu_r, \sigma_r^2) 
\end{align*}
where $S_p$ and $Z_r$ are normalization constants and
\begin{align*}
\sigma_p^2 &= \left(\frac{1}{\sigma_1^2} + \frac{1}{\sigma_2^2}\right)^{-1} \\
\mu_p &= \sigma_p^2 \cdot \left(\frac{\mu_1}{\sigma_1^2} + \frac{\mu_2}{\sigma_2^2}\right) \\
\sigma_r^2 &= \left(\frac{1}{\sigma_1^2} - \frac{1}{\sigma_2^2}\right)^{-1} \\
\mu_r &= \sigma_r^2 \cdot \left(\frac{\mu_1}{\sigma_1^2} - \frac{\mu_2}{\sigma_2^2}\right)
\end{align*}
with $\sigma_1 < \sigma_2$ in the latter case (ratio).
It is easy to observe that using the natural parameterization of the Gaussian distribution with sufficient statistics given by
\begin{align*}
\chi = \frac{\mu}{\sigma^2} \\
\xi = \frac{1}{\sigma^2},
\end{align*}
products and ratios of Gaussians translates respectively into the sum and difference of the natural parameters $\chi$ and $\xi$. The implementation used in the code uses natural parameter Gaussian distributions, and sum and difference of its parameters to obtain the required products and ratios. This parameterization and the consequences on product and ratios easily generalize to any exponential family distribution, but it is here presented in the Gaussian case for convenience.

\section{Omitted proofs}
\begin{proof}[Proof of \Cref{prop:ep_free_energy}]
At step $t$ the global posterior for server parameters is $s^{(t)}(\bm{\theta}) = s_{i}(\bm{\theta})\prod_{j \ne i}^K s_j^{(t-1)}(\bm{\theta})$ and analogously the client parameters distribution reads $c^{(t)}(\bm{\phi}) = c_{i}(\bm{\phi}_i)\prod_{j\ne i}^K c_j^{(t-1)}(\bm{\phi}_j)$. Then the {\tt EP}-like update for the model described is given by minimizing the following KL divergence w.r.t $ s_{i}(\bm{\theta})$ and $c_{i}(\bm{\phi}_{i})$
\begin{align*}
D_{KL}&\left(s^{(t)}(\bm{\theta}) c^{(t)}(\bm{\phi}) \bigg|\bigg| \frac{s^{(t)}(\bm{\theta}) c^{(t)}(\bm{\phi})}{s_{i}(\bm{\theta}) c_{i} (\bm{\phi}_i)} \frac{p(\bm{\theta},\bm{\phi}_i | \mathcal{D}_i)}{p(\bm{\theta})^{\frac{k-1}{k}} }  \right) = \\
&= \int d \bm{\theta} \ s^{(t)}(\bm{\theta}) \log s_{i}(\bm{\theta}) \int d \bm{\phi} \ c^{(t)}(\bm{\phi}) \\
&\hspace{1cm} + \int d \bm{\theta}\ s^{(t)}(\bm{\theta}) \int d \bm{\phi}\ c^{(t)}(\bm{\phi}) \log c_{i}(\bm{\phi}_i) \\
&\hspace{1cm}- \int d \bm{\theta}\ d \bm{\phi}\ s^{(t)}(\bm{\theta}) c^{(t)}(\bm{\phi}) \log \frac{p(\bm{\theta},\bm{\phi}_i | \mathcal{D}_i)}{p(\bm{\theta})^{\frac{k-1}{k}} } \\
&= \int d \bm{\theta}\ s^{(t)}(\bm{\theta}) \log \frac{s_{i}(\bm{\theta}) s^{(t)}(\bm{\theta}) }{s^{(t)}(\bm{\theta}) p(\bm{\theta})^{\frac{1}{K}}} \\
&\hspace{0.3cm} + \int d \bm{\phi}_i\ c^{(t)}_i(\bm{\phi}_i) \log \frac{c_{i}^{(t)}(\bm{\phi}_i)}{p(\bm{\phi}_i)} + \\
&\hspace{0.3cm} - \int d \bm{\theta}\ s^{(t)}(\bm{\theta}) \int d \bm{\phi}_i\ c^{(t)}_i(\bm{\phi}_i) \log p(\mathcal{D}_i|\bm{\theta},\bm{\phi}_i)
\end{align*}
where the second equality comes from the normalization of client and server pdfs and from Bayes rule $p(\bm{\theta},\bm{\phi}_i|\mathcal{D}_i) \propto p(\mathcal{D}_i|\bm{\theta},\bm{\phi}_i) p(\bm{\phi}_i) p(\bm{\theta})$. Notice also that $\frac{s_{i}(\bm{\theta})} {s^{(t)}(\bm{\theta})} = \frac{s_{i}^{(t-1)}(\bm{\theta})}{s^{(t-1)}(\bm{\theta})}$ because of the factorization in \Cref{eq:factorization}, and hence \Cref{eq:free_energy} is proved.
\end{proof}

\section{Full experiment details}
In all experiments reported here and in the main text we use a batch size $B = 20$, except for the Shakespeare dataset, where $B=10$. For both baselines and Virtual we perform a careful log spaced grid-search over 5 values of the regularization multiplier $\beta$ and of the \emph{client} learning rate $\eta_c$, verifying that optimal values do not lie at the boundaries of the grid. Similarly we use a linearly spaced grid for the \emph{server} learning rate $\eta_s$ of the type $\{0.2, 0.4 \dots, 1.\}$. In the case of Virtual, we use an additional damping factor $\gamma \in [0, 1]$, frequently used in the case of message passing algorithms (see \cite{minka2001expectation, minka2004power, bui2018partitioned}), to prevent oscillations. The damping factor acts on the client updates as $s_i(\theta)_{t+1} \leftarrow s_i(\theta)_{t+1}^\gamma s_i(\theta)_{t}^{1-\gamma}$. In order to retain the same number of hyper-parameters as the baselines FedProx and FedAvg we fix the value of the damping factor $\gamma$ as $1 - \eta_s$ that resulted in good performance of the overall method. 
%Full details are available in the configuration files of the code given in the supplementary material.

\newpage
\bibliography{bib}
\bibliographystyle{plain}

\end{document}